\documentclass[10pt,twocolumn,letterpaper]{article}

\usepackage{cvpr}              %
\usepackage[accsupp]{axessibility}  %

\usepackage{graphicx}
\usepackage{amsmath}
\usepackage{amssymb}
\usepackage{mathtools}
\usepackage{booktabs}
\usepackage{color}
\usepackage{xcolor} %
\usepackage{algorithm}
\usepackage{algorithmic}
\usepackage{xspace}
\usepackage{multirow}
\usepackage{array}
\usepackage{wrapfig}

\usepackage[pagebackref,breaklinks,colorlinks]{hyperref}

\definecolor{myred}{RGB}{200, 1, 80}
\newcolumntype{P}[1]{>{\centering\arraybackslash}p{#1}}

\newcommand{\deunsol}[1]{\textcolor{black}{#1}}
\newcommand{\ds}[1]{\textcolor{black}{#1}}

\newcommand{\ours}{SQUAT\xspace}

\usepackage[capitalize]{cleveref}
\crefname{section}{Sec.}{Secs.}
\Crefname{section}{Section}{Sections}
\Crefname{table}{Table}{Tables}
\crefname{table}{Tab.}{Tabs.}

\def\ie{\emph{i.e.}}
\def\eg{\emph{e.g.}}

\def\cvprPaperID{4074} %
\def\confName{CVPR}
\def\confYear{2023}

\begin{document}

\title{ Devil's on the Edges: Selective Quad Attention for Scene Graph Generation}

\author{Deunsol Jung \hspace{0.8cm}  Sanghyun Kim \hspace{0.8cm} Won Hwa Kim \hspace{0.8cm} Minsu Cho \vspace{2.0mm}\\
Pohang University of Science and Technology (POSTECH), South Korea \\
\small
\href{http://cvlab.postech.ac.kr/research/SQUAT}{\url{http://cvlab.postech.ac.kr/research/SQUAT}}
}
\maketitle

\begin{abstract}

Scene graph generation aims to construct a semantic graph structure from an image such that its nodes and edges respectively represent objects and their relationships. 
One of the major challenges for the task lies in the presence of distracting objects and relationships in images; contextual reasoning is strongly distracted by irrelevant objects or backgrounds and, more importantly, a vast number of irrelevant candidate relations.   
To tackle the issue, we propose the Selective Quad Attention Network (\ours) that learns to select relevant object pairs and disambiguate them via diverse contextual interactions. 
\ours consists of two main components: edge selection and quad attention. 
The edge selection module selects relevant object pairs, i.e., edges in the scene graph, which helps contextual reasoning, and the quad attention module then updates the edge features using both edge-to-node and edge-to-edge cross-attentions to capture contextual information between objects and object pairs. 
Experiments  demonstrate the strong performance and robustness of \ours, achieving the state of the art on the Visual Genome and Open Images v6 benchmarks. 
\end{abstract}
\section{Introduction}

The task of scene graph generation (SGG) is to construct a visually-grounded graph from an image such that its nodes and edges respectively represent objects and their relationships in the image~\cite{lu2016visual,yang2018graph,xu2017scene}. 
The scene graph provides a semantic structure of images beyond individual objects and thus is useful for a wide range of vision problems such as visual question answering~\cite{teney2017graph,tang2019learning}, image captioning~\cite{zhong2020comprehensive}, image retrieval~\cite{johnson2015image}, and conditional image generation~\cite{johnson2018image}, where a holistic understanding of the relationships among objects is required for high-level reasoning.  
With recent advances in deep neural networks for visual recognition, SGG has been actively investigated in the computer vision community.
A vast majority of existing methods tackle SGG by first detecting candidate objects and then performing contextual reasoning between the objects via message passing~\cite{li2021bipartite,xu2017scene,li2017scene} or sequential modeling~\cite{lu2021context,tang2019learning,zellers2018neural}. 
Despite these efforts, the task of SGG  remains extremely challenging, and even the state-of-the-art methods do not produce reliable results for practical usage. 
\begin{figure}[t!] %
\begin{center}
\includegraphics[width=1.0\linewidth]{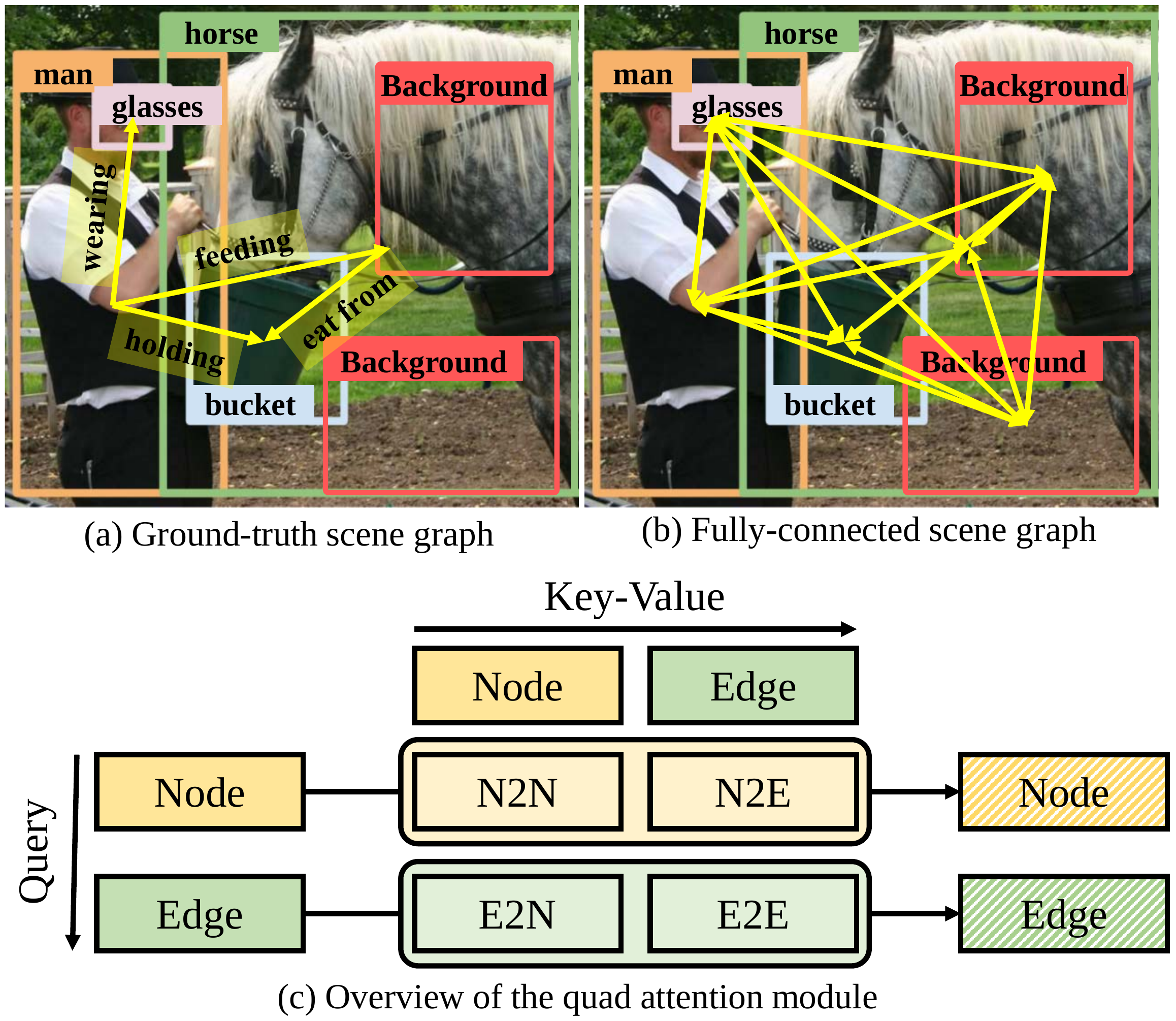}
\end{center}
\caption{(a) The ground-truth scene graph contains only 4 ground-truth objects and 4 relations between the objects. (b) Only 13\% of edges in a fully-connected graph have the actual relationships according to the ground-truths. (c) The overview of the quad attention. The node features are updated by node-to-node (N2N) and node-to-edge (N2E) attentions, and the edge features are updated by edge-to-node (E2N) and edge-to-edge (E2E) attentions. }
\label{fig:teaser}
\end{figure}

While there exist a multitude of challenges for SGG, the intrinsic difficulty may lie in the presence of distracting objects and relationships in images;  
there is a vast number of potential but irrelevant relations, \ie, edges, which quadratically increase with the number of candidate objects, \ie, nodes, in the scene graph.
The contextual reasoning for SGG in the wild  is thus largely distracted by irrelevant objects and their relationship pairs. 
Let us take a simple example as in Fig.~\ref{fig:teaser}, where 4 objects and 4 relations in its ground-truth scene graph exist in the given image. 
If our object detector obtains 6 candidate boxes, 2 of which are from the background (red), then the contextual reasoning, \eg, message passing or self-attention, needs to consider 30 potential relations, 87\% of which are not directly related according to the ground-truth and most of them may thus act as distracting outliers. 
In practice, the situation is far worse; in the Visual Genome dataset, the standard benchmark for SGG, an image contains 38 objects and 22 relationships on average~\cite{xu2017scene}, which means that only around 1\% of object pairs 
have direct and meaningful relations even when  object detection is perfect. 
As will be discussed in our experiments, we find that existing contextual reasoning schemes obtain only a marginal gain at best and often degrade the performance. 
The crux of the matter for SGG may lie in developing a robust model for contextual reasoning against irrelevant objects and relations. 

To tackle the issue, we propose the {\it Selective Quad Attention Network (\ours)} that learns to select relevant object pairs and disambiguate them via diverse contextual interactions with objects and object pairs.   
The proposed method consists of two main components: edge selection and \deunsol{quad} attention. 
The edge selection module removes irrelevant object pairs, which may distract contextual reasoning, by predicting the relevance score for each pair. 
The quad attention module then updates the edge features using edge-to-node and edge-to-edge cross-attentions as well as the node features using node-to-node and node-to-edge cross-attentions; it thus captures contextual information between all objects and object pairs, as shown in Figure~\ref{fig:teaser} (c).  
Compared to previous methods~\cite{li2017scene,li2021bipartite}, which perform either node-to-node or node-to-edge interactions, 
our \deunsol{quad} attention provides more effective contextual reasoning by capturing diverse interactions in the scene graph. 
For example, in the case of Fig.~\ref{fig:teaser} (a), [`man', `feeding', `horse'] relates to [`man', `holding', `bracket'] and [`horse', `eat from', `bracket'], and vice versa; node-to-node or node-to-edge interactions are limited in capturing such relations between the edges.

Our contributions can be summarized as follows: %
\begin{itemize}
    \item We introduce the edge selection module for SGG that learns to select relevant edges for contextual reasoning. 
    \item We propose the quad attention module for SGG that performs effective contextual reasoning by updating node and edge features via diverse interactions.
    \item The proposed SGG model, \ours, outperforms the state-of-the-art methods on both Visual Genome and Open Images v6 benchmarks. In particular, \ours achieves remarkable improvement on the SGDet settings, which is the most realistic and challenging. 
\end{itemize}

\section{Related work}
\paragraph{Scene graph generation}
The vast majority of SGG methods~\cite{li2021bipartite,zellers2018neural,li2017scene} predict scene graphs in two stages: object detection and contextual reasoning. 
While the first stage is typically done by a pre-trained detection module~\cite{ren2015faster,carion2020end}, 
contextual reasoning is performed by different types of message passing~\cite{li2021bipartite,lin2020gps,xu2017scene,khandelwal2021segmentation,zareian2020bridging,woo2018linknet,li2018factorizable,chen2019knowledge,li2017scene,dai2017detecting,wang2019exploring,qi2019attentive,yin2018zoom}, which uses a graph neural network with node-to-edge and edge-to-node attentions, or sequential modeling~\cite{lu2021context,tang2019learning,zellers2018neural,gu2019scene}, which updates the node features with node-to-node attention and constructs edge features with edge-to-node attention. 
Unlike the previous methods, we propose quad attention, which comprises node-to-node, node-to-edge, edge-to-node, and edge-to-edge interactions, to capture all types of context exchange between candidate objects and their pairs for relational reasoning. 
\deunsol{In contextual reasoning, most of the methods consider all the candidate object pairs, \ie, a fully-connected graph whose nodes are candidate objects.}
While Graph R-CNN~\cite{yang2018graph} proposes a relation proposal network that prunes the edges from a fully-connected graph, it focuses on reducing the cost of message passing and does not analyze the effect of edge selection on the performance of scene graph generation. In contrast, we introduce an effective edge selection method and provide an in-depth analysis of it.   
On the other hand, since dataset imbalance/bias has recently emerged as a critical bottleneck for learning SGG\footnote{For example, in the Visual Genome dataset, the most frequent entity class is 35 times larger than the least frequent one, and the most frequent predicate class is 8,000 times larger than the least frequent one.},
several methods~\cite{suhail2021energy,Yao_2021_ICCV,Knyazev_2021_ICCV,chiou2021recovering,wang2020tackling} propose to adopt the techniques from long-tailed recognition, \eg, data resampling~\cite{li2021bipartite,Desai_2021_ICCV} and loss reweighting~\cite{Guo_2021_ICCV,knyazev2020graph,yan2020pcpl}.

\begin{figure*}[t!]
\centering
\includegraphics[width=0.99\linewidth]{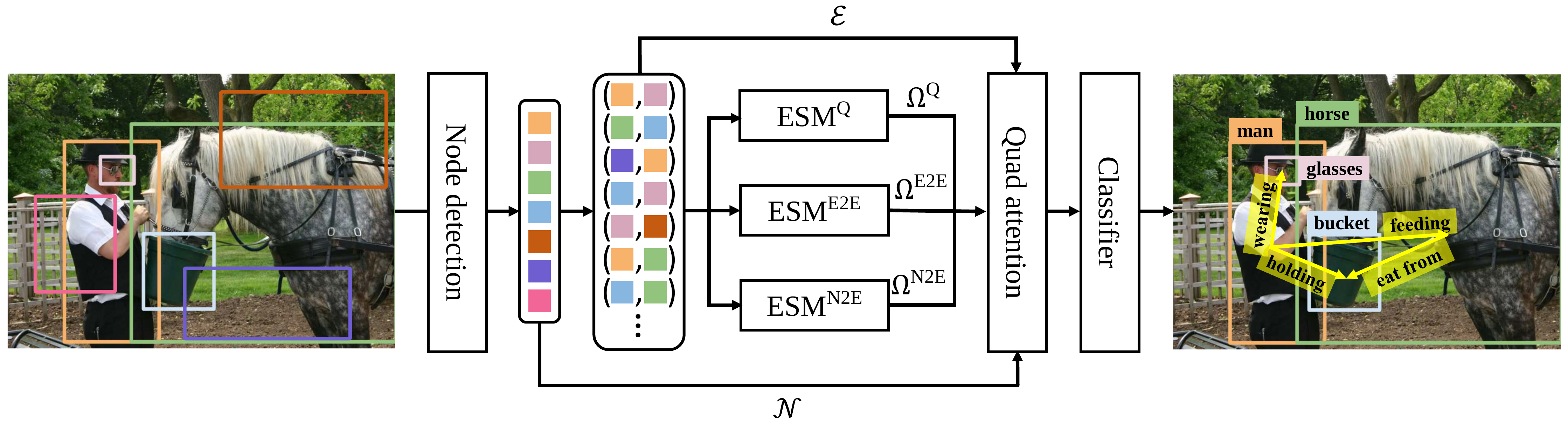}
\caption{
The overall architecture of Selective Quad Attention Networks (\ours). \ours consists of three components: the node detection module, the edge selection module, and the quad attention module. 
First, the node detection module extracts nodes $\mathcal{N}$ by detecting object candidate boxes and extracting their features. 
Also, all possible pairs of the nodes are constructed as initial edges $\mathcal{E}$. 
Second, the edge selection module select valid edges $(\Omega^{\mathrm{Q}}, \Omega^{\mathrm{E2E}}, \Omega^{\mathrm{N2E}})$ with high relatedness scores. 
Third, the quad attention module updates the node and edge features via four types of attention. 
Finally, the output features are passed into a classifier to predict the scene graph. 
See Sec.~\ref{sec:arch} for the details.  }
\label{fig:overall}
\end{figure*}

\paragraph{Transformers for vision tasks and graph structures}

Transformers~\cite{vaswani2017attention} have been adapted to the various computer vision tasks, \eg, object classification~\cite{dosovitskiy2020image,liu2021swin}, object detection~\cite{carion2020end,zhu2020deformable,roh2021sparse,liu2021swin} and segmentation~\cite{zheng2021rethinking,liu2021swin}, and also extended for graph structures~~\cite{kim2021transformers,min2022transformer,rong2020self,lin2021mesh}. 
Despite their success, %
vision transformer networks typically suffer from high complexity and memory consumption. 
Several variants of transformer networks~\cite{wang2020linformer,choromanski2020rethinking,zhu2020deformable,kitaev2020reformer,roh2021sparse} have been proposed to tackle the issue and showed that a proper sparsification technique, \eg, Sparse DETR~\cite{roh2021sparse}, can not only reduce the cost of computation and memory but also improve the task performance.
Our transformer network is designed to perform contextual reasoning for scene graph generation by capturing the inherent relationships between objects and relevant object pairs, and 
unlike existing sparsification methods, which focus on token pruning~\cite{roh2021sparse} or local attention~\cite{zhu2020deformable,kitaev2020reformer}, our edge selection module prunes not only the query edges to update but also the key-value edges used for updating.

\section{Problem Definition}
\label{sec:problem}
Given an image $I$, the goal of SGG is to generate a visually grounded graph $G=(\mathcal{O},\mathcal{R})$ that represents objects $\mathcal{O}$ and their semantic relationships $\mathcal{R}$ for object classes $\mathcal{C}$ and predicate classes $\mathcal{P}$. 
An object  
$o_i \in \mathcal{O} $ is described by a pair of a bounding box
$b_i \in [0,1]^4$ and its class label $c_i \in \mathcal{C}$: $o_i = (b_i, c_i)$.
A relationship $r_k \in \mathcal{R}$ is represented by a triplet of a subject $o_i\in \mathcal{O}$, an object $o_j\in \mathcal{O}$, and a predicate label $p_{ij} \in \mathcal{P}$: $r_k = (o_i, o_j, p_{ij})$, which represents relationship $p_{ij}$ between  subject $o_i$ and object $o_j$.

\section{Selective Quad Attention Networks}
\label{sec:arch}
To generate semantically meaningful scene graphs as described in Section \ref{sec:problem}, we propose the Selective Quad Attention Network (\ours) that consists of three main components as shown in Fig.~\ref{fig:overall}: the node detection module (Sec.~\ref{sec:object_detector}), the edge selection module (Sec.~\ref{sec:pair_gating}), and the quad attention module (Sec.~\ref{sec:p2p_decoder}).
First, the node detection module establishes nodes for a scene graph by detecting object candidate boxes and extracting their features. All possible pairs of the nodes are constructed as potential edges.  
Second, among all the potential edges, the edge selection module selects valid edges with high relatedness scores. 
Third, the quad attention module updates the features of nodes and valid edges via four types of attention: node-to-node (N2N), node-to-edge (N2E), edge-to-node (E2N), and edge-to-edge (E2E). 
For the quad attention module, we use three edge selection modules: query edge selection module for entire quad attention ($\mathrm{ESM}^Q$) and key-value edge selection modules for N2E attention ($\mathrm{ESM}^{\mathrm{N2E}}$) and E2E attention (${\mathrm{ESM}}^{\mathrm{E2E}}$). 
The nodes and edges may require different sets of edges to update their features, and some pruned edges may help to update nodes or selected edges. 
For example, an edge between a person and a background, \eg, an ocean, is invalid but can help to predict the relationships between a person and other objects.
Only the valid edges extracted by $\mathrm{ESM}^{\mathrm{N2E}}$ and ${\mathrm{ESM}}^{\mathrm{E2E}}$ are used to update the features of the nodes and valid edges from $\mathrm{ESM}^Q$.
Finally, the output features are then passed into a classifier %
to predict relationship classes. 
The remainder of this section presents the details of each component and training procedure (Sec.~\ref{sec:training}). 
In this section, the calligraphic font, \ie, $\mathcal{N}$ and $\mathcal{E}$, denotes a set of features while the italic, \ie, $N$ and $E$, denotes a matrix of stacked features of the set. 

\subsection{Node detection for object candidates}
\label{sec:object_detector}

Given an image $I$, we use a pre-trained object detector, \ie, Faster R-CNN~\cite{ren2015faster} in our experiments, to extract object bounding boxes and their class labels.
Let $b_i\in[0,1]^4$ be the $i$-th object box coordinate and $v_i\in\mathbb{R}^{d_v}$ its visual feature %
where $d_v$ is the dimension of the visual feature. 
We construct a node feature $f_i$ by transforming $b_i$ and $v_i$ via   
\begin{equation}
    f_i = W_o[W_v v_i;W_g b_i],
\end{equation}
where $W_o$, $W_v$, and $W_g$ are linear transformation matrices and $[\cdot;\cdot]$ is a concatenation operation. 
The edge feature $f_{ij}$ is formed by concatenating two node features $f_i$ and $f_j$ and performing a linear transformation as 
\begin{equation}
    f_{ij} = W_p[f_i;f_j],
\label{eq:feature_extraction}
\end{equation}
where $W_p$ is the linear transformation matrix. As in Fig.~\ref{fig:overall}, the set of entire node features $\mathcal{N}=\{f_i|1 \leq i \leq n\}$ and the set of all possible edge features $\mathcal{E}=\{f_{ij}|1\leq i, j\leq n, i \neq j\}$ are passed into the edge selection and \deunsol{quad} attention modules, whose details are described below. 
\deunsol{We denote the stacks of the features in $\mathcal{N}$ and $\mathcal{E}$ as $N$ and $E$ for the sake of simplicity. }

\begin{figure}[t!]
\centering
\includegraphics[width=0.99\linewidth]{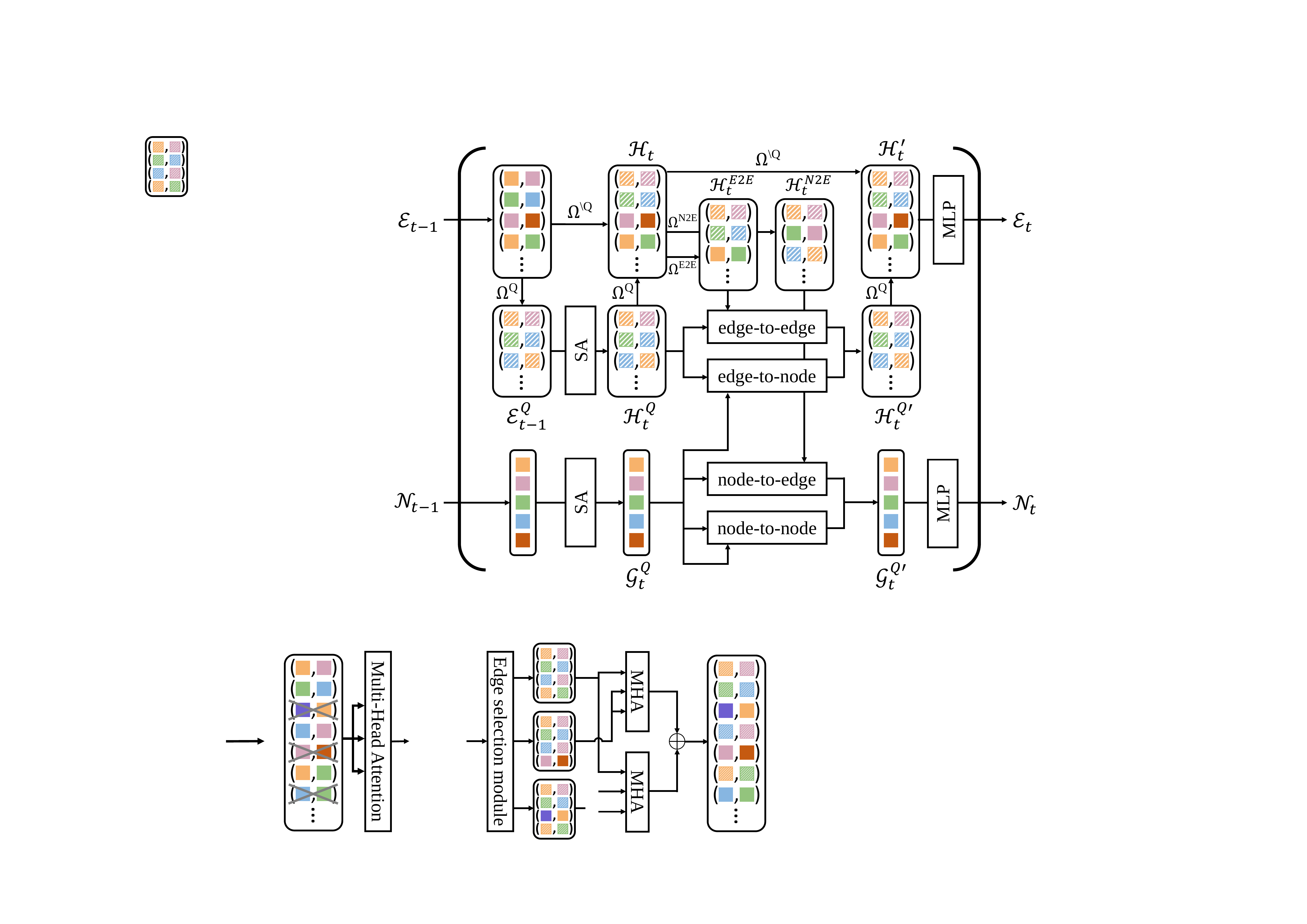}
\caption{Detailed architecture of the quad attention. The node features are updated by node-to-node and node-to-edge attentions, and the valid edge features, selected by $\mathrm{ESM}^\mathrm{Q}$, are updated by edge-to-node and edge-to-edge attentions. The key-value of node-to-edge and edge-to-edge attentions are selected by $\mathrm{ESM}^\mathrm{N2E}$ and $\mathrm{ESM}^\mathrm{E2E}$. See Sec.~\ref{sec:p2p_decoder} for the details. Best viewed in color.  }
\label{fig:quad_attention}
\end{figure}

\subsection{Edge selection for relevant object pairs}
\label{sec:pair_gating}
While node features $N$ and edge features $E$ can be updated via attentive message passing, 
a large number of irrelevant edges in $E$  interferes with the attention process. %
We thus propose to prune invalid edges (\ie, non-existing/false relationships) before proceeding to the \deunsol{quad} attention module, which will be described in the next subsection. 

In order to remove such distracting edges, we introduce an edge selection module (ESM) that takes an edge feature $f_{ij}$ between nodes $i$ and $j$ and predicts its relatedness score $s_{ij}$ using a \deunsol{simple multi-layer perceptron}.
We choose the pairs with top-$\rho\%$ highest relatedness scores as valid edges to use in the following quad attention module.

As mentioned \deunsol{earlier}, we use three edge selection modules: $\mathrm{ESM}^\mathrm{Q}$, $\mathrm{ESM}^\mathrm{N2E}$, and $\mathrm{ESM}^\mathrm{E2E}$. 
Each edge selection module $\mathrm{ESM}^a$ takes the initial edge features $\mathcal{E}$ as inputs and outputs the valid edge index set $\Omega$ for each module, resulting in $\Omega^\mathrm{Q}$, $\Omega^\mathrm{E2E}$, and $\Omega^\mathrm{N2E}$.

\subsection{Quad attention for relationship prediction}
\label{sec:p2p_decoder}
To capture %
contextual interactions between the nodes and the edges, 
we propose a \deunsol{quad} attention scheme inspired by the transformer decoder~\cite{vaswani2017attention}. 
The main component of the \deunsol{quad} attention is multi-head attention: 
\begin{align}
    &\mathrm{MHA}(Q,K,V) = [\mathrm{HA}_1; \cdots; \mathrm{HA}_h]W^O \\
    &\mathrm{HA}_i = \mathrm{softmax}\left(\frac{(QW_i^\mathrm{Q})(KW_i^\mathrm{K})^T}{\sqrt{d_k}} \right)VW_i^\mathrm{V},
    \label{eq:self-attention}
\end{align}
where $Q, K$, and $V$ are query, key, and value matrices.
$W_i^\mathrm{Q}, W_i^\mathrm{K}$, and $W_i^\mathrm{V}$ are learnable transformation parameters for $Q, K$, and $V$, respectively, 
$d_k$ is the dimension of the query vector, and 
$W^\mathrm{O}$ is a learnable transformation parameter for the output. 
Each attention head $\mathrm{HA}_i$ captures the information from different representation subspaces in parallel, and the multi-head attention aggregates them.

Fig.~\ref{fig:quad_attention} shows the architecture of our quad attention layer.
Following the transformer decoder layer, the $t$-th \deunsol{quad} attention layer takes output edge features $E_{t-1}$ and node features $N_{t-1}$ from the $(t-1)$-th layer as its input and update them with a self-attention first. 
Instead of updating all possible edge features $E_{t-1}$, we only update the valid edge features $E_{t-1}^\mathrm{Q}$, whose indices are in $\Omega^\mathrm{Q}$ extracted from $\mathrm{ESM}^\mathrm{Q}$:
\begin{align}
    G_{t} &= \mathrm{LN}(N_{t-1} + \mathrm{MHA}(N_{t-1}, N_{t-1}, N_{t-1})), \\
    H_{t}^{\mathrm{Q}} &= \mathrm{LN}(E_{t-1}^{\mathrm{Q}} + \mathrm{MHA}(E_{t-1}^{\mathrm{Q}}, E_{t-1}^{\mathrm{Q}}, E_{t-1}^{\mathrm{Q}})),
\end{align}
where $\mathrm{LN}$ is layer normalization, $G_t$ and $H_t$ are the output of the self-attention layer for node features and valid edge features, respectively. 
For key-value edge features of N2E and E2E attentions, we extract the key-value set from the updated entire edge set $\mathcal{H}_t=\mathcal{H}^\mathrm{Q}_t \cup \mathcal{E}^{\backslash \mathrm{Q}}$, where $\mathcal{H}_t^\mathrm{Q}$ is the set of updated valid edges for query and $\mathcal{E}^{\backslash \mathrm{Q}}=\mathcal{E} - \mathcal{E}^\mathrm{Q}$. 
Then, $H_t^\mathrm{Q}$ are refined by E2N and E2E attentions and $G_t$ are refined by N2N and N2E attentions:
\begin{align}
    \phantom{H^{\mathrm{Q}\prime}_t} & \begin{aligned}
    \mathllap{G^{\prime}_{t}} &= \mathrm{LN}(G_t + \underbrace{\mathrm{MHA}(G_t, G_t, G_t)}_{\text{node-to-node attention}} \\ 
    & \qquad\qquad\quad  + \underbrace{\mathrm{MHA}(G_t,  H_t^{\mathrm{N2E}},  H_t^{\mathrm{N2E}}))}_{\text{node-to-edge attention}},  
    \end{aligned}
\end{align}
\begin{align}
   & \begin{aligned}
    \mathllap{H^{\mathrm{Q}\prime}_t} &=\mathrm{LN}(H^{\mathrm{Q}}_{t}+\underbrace{\mathrm{MHA}(H^{\mathrm{Q}}_{t}, G_{t}, G_{t})}_{\text{edge-to-node attention}} \\
    & \qquad\qquad  +\underbrace{\mathrm{MHA}(H^{\mathrm{Q}}_{t}, H_t^{\mathrm{E2E}}, H_t^{\mathrm{E2E}}))}_{\text{edge-to-edge attention}},
    \end{aligned}
\end{align}
where $H_t^{\mathrm{N2E}}$ and $H_t^{\mathrm{E2E}}$ are selected by the indices $\Omega^\mathrm{N2E}$ and $\Omega^\mathrm{E2E}$ from the stack of $\mathcal{H}_t$, \ie, $H_t$. 
Each attention explicitly represents a particular type of relationship between edges and nodes and helps to construct contextual information for the scene graph generation. 
\ds{Lastly, $G_t'$ and $H_t'$ are further updated by multi-layer perceptron (MLP) followed by the residual connection and a layer normalization: 
\begin{align}
    \label{eq:lin_node}
    N_t&=\mathrm{LN}(G_t' + \mathrm{MLP}(G_t')) \\
    E_t&=\mathrm{LN}(H_t' + \mathrm{MLP}(H_t')),
    \label{eq:lin_edge}
\end{align}
where $H_t'$ is the stack of $\mathcal{H}_t'=\mathcal{H}^{\mathrm{Q}\prime} \cup \mathcal{E}^{\backslash \mathrm{Q}}$, and the quad attention layer outputs $N_t$ and $E_t$. }

The inputs $N_0$ and $E_0$ of the first \deunsol{quad} attention layer are the entire node features $N$ and all possible edge features $E$, which are defined in Sec.~\ref{sec:object_detector}. 
Every quad attention layer uses the same valid edge sets \deunsol{to update the node features and valid edge features by four types of attention}. 
Given the output edge features $E_T$ of the last $T$-th quad attention layer, each edge feature $e_{ij} \in \mathcal{E}_T$ is passed into a feedforward MLP to produce a probability distribution $y_{ij}$ over the predicate classes $\mathcal{P}$.

\subsection{Training objective}
\label{sec:training}

To train \ours, we use %
a combination of two loss functions: a cross-entropy loss for the predicate classification and a binary cross-entropy loss for the edge selection module. 
The first predicate classification loss is defined as:
\begin{equation}
    \mathcal{L}_{\mathrm{PCE}}=\frac{1}{|\mathcal{E}|}\sum^{|\mathcal{N}|}_{i,j=1, i\neq j}\mathcal{L}_{\mathrm{CE}}(y_{ij}, \hat{y}_{ij}),
\end{equation}
\ds{where $\mathcal{L}_{\mathrm{CE}}$ is the cross-entropy loss and $\hat{y}_{ij}$ is a one-hot vector of ground-truth relationship labels $\hat{p}_{ij}$ between object $i$ and object $j$. }
To train the edge selection module, we use auxiliary binary cross-entropy defined as: 
\begin{equation}
    \mathcal{L}_{\mathrm{ESM}}^a=\frac{1}{|\mathcal{E}|}
    \sum^{|\mathcal{N}|}_{i,j=1, i\neq j}\mathcal{L}_{\mathrm{BCE}}(s_{ij}^a, \hat{s}_{ij}),
\end{equation}
where $\mathcal{L}_{\mathrm{BCE}}$ is the binary cross-entropy loss, $\hat{s}_{ij}$ is the binary indicator of whether object $i$ and object $j$ have a relationship or not, and $a \in\mathcal{A}=\{\mathrm{Q}, \mathrm{E2E}, \mathrm{N2E}\}$. The entire loss is defined as:
\begin{equation}
    \mathcal{L}=\mathcal{L}_{\mathrm{PCE}}+\lambda \frac{1}{|\mathcal{A}|}\sum_{a\in\mathcal{A}}\mathcal{L}_{\mathrm{ESM}}^a, 
\end{equation}
where $\lambda > 0$ is a hyper-parameter. 
In training, $\mathcal{L}_{\mathrm{CE}}$ does not affect the parameters of ESM directly due to the hard selection of ESM, and the gradient passes on to train the edge feature extraction; ESM is mainly trained by $\mathcal{L}_\mathrm{ESM}$.
\section{Experiments}
In this section, we perform a diverse set of experiments to evaluate the proposed model. 
We use two datasets: 1) Visual Genome (VG)~\cite{krishnavisualgenome} and 2) OpenImages v6~\cite{OpenImages2} datasets 
to train and evaluate model performances. 
We intend to show that our model can be generalized over heterogeneous cases 
by demonstrating competitive results on the two independent datasets. 

\begin{table*}[!t]
\begin{center}
\scalebox{0.90}{
\begin{tabular}{l|P{1.3cm}P{1.3cm}P{1.3cm}|P{1.3cm}P{1.3cm}P{1.3cm}|P{1.3cm}P{1.3cm}P{1.3cm}}
\toprule
\multirow{2}{*}{Methods} & \multicolumn{3}{c|}{PredCls} & \multicolumn{3}{c|}{SGCls} & \multicolumn{3}{c}{SGDet} \\ 

 & mR@20 & mR@50 & mR@100 & mR@20 & mR@50 & mR@100 & mR@20 & mR@50 & mR@100  \\ \midrule
IMP+$^\ddagger$ ~\cite{xu2017scene} & 8.9&11.0&11.8&5.2&6.2&6.5&2.8&4.2&5.3 \\
Motifs$^\ddagger$ ~\cite{zellers2018neural} & 11.5&14.6&15.8&6.5&8.0&8.5&4.1&5.5&6.8 \\
RelDN~\cite{zhang2019graphical}& - & 15.8 & 17.2 & - & 9.3& 9.6 & - & 6.0& 7.3\\
VCTree$^\ddagger$~\cite{tang2019learning}& 12.4& 15.4& 16.6& 6.3& 7.5& 8.0& 4.9&6.6&7.7 \\
MSDN~\cite{li2017scene}& - & 15.9 & 17.5 & - & 9.3 & 9.7 & 6.1 & 7.2 \\
GPS-Net~\cite{lin2020gps}&-&15.2&16.6&-&8.5&9.1&-&6.7&8.6 \\ 
RU-Net~\cite{lin2022ru} & - & - & 24.2 & - & - & 14.6 & - & - & 10.8 \\
HL-Net~\cite{lin2022hl} & - & - & 22.8 & - & - & 13.5 & - & - & 9.2 \\
VCTree-TDE~\cite{tang2020unbiased}& 18.4 & 25.4 & 28.7& 8.9 & 12.2 & 14.0 & 6.9&9.3 & 11.1 \\
Seq2Seq~\cite{lu2021context}& 21.3 & 26.1 & 30.5& 11.9 & 14.7& 16.2 & 7.5 & 9.6& 12.1 \\  
GPS-Net$^\dagger$& 21.5 & 27.1 & 29.1 & 6.4 & 10.1 & 12.3 & 6.6 & 9.4 & 11.9 \\
JMSGG~\cite{xu2021joint}& - & 24.9 & 28.0 & - &13.1 & 14.7 &  - &9.8 & 11.8  \\
BGNN$^\dagger$~\cite{li2021bipartite} & - & 30.4& 32.9& -&14.3&16.5&-&10.7&12.6 \\ 
\ours$^\dagger$ (Ours) & \bfseries{25.6} & \bfseries{30.9} & \bfseries{33.4} & \bfseries{14.4}   & \bfseries{17.5}  & \bfseries{18.8}  &   \bfseries{10.6}   &     \bfseries 14.1   &   \bfseries  16.5  \\
\bottomrule
\end{tabular}
}
\caption{The scene graph generation performance of three subtasks on Visual Genome (VG) dataset with graph constraints. $\dagger$ denotes that the bi-level sampling~\cite{li2021bipartite} is applied for the model. $\ddagger$ denotes that the results are reported from the \cite{tang2020unbiased}.
\label{tb:maintable}}
\end{center}
\end{table*}

\subsection{Datasets and evaluation metrics}
\subsubsection{Visual Genome~\cite{krishnavisualgenome}} 
The Visual Genome dataset is composed of 108k images with an average of 38 objects and 22 relationships per image. 
However, most of the predicate classes have less than 10 samples. 
Therefore, we adopt the widely-used VG split~\cite{li2017scene,zellers2018neural} to select the most frequent 150 object classes and 50 predicate classes. 
Following the \cite{zellers2018neural}, 
we first split the dataset into a training set~($70\%$) and a test set~($30\%$). 
Then, we sample 5k validation images from the training set to tune the hyperparameters.
We evaluate \ours on three subtasks: Predicate Classification (PredCls), Scene Graph Classification (SGCls), and Scene Graph Detection (SGDet). 
The PredCls predicts the relationships given the ground-truth bounding boxes and object labels, 
the SGCls aims to predict the object labels and the relationships given the ground-truth bounding boxes, and 
the SGDet targets predicting the object bounding boxes, object labels, and relationships without any ground-truth. 
As the evaluation metrics, we adopt the mean recall@K (mR@$K$), as previously used in scene graph generation literature~\cite{chen2019knowledge,tang2020unbiased}. 
mR@$K$ is the average of recall@$K$ for each relation.
Following~\cite{xu2017scene}, we apply the graph-constraint, in which each object pair can have only one relationship, for evaluation.

\vspace{-1mm}
\subsubsection{OpenImages v6~\cite{OpenImages2}} 
The OpenImages v6 dataset has 126,368 images for the training, 1,813 images for the validation, and 5,322 images for the test.  
Each image in the dataset has 4.1 objects and 2.8 relationships on average. 
The dataset has 301 object classes and 31 predicate classes. 
Compared with the Visual Genome dataset, the quality of annotation is far more robust and complete. %
For OpenImages v6, following \cite{OpenImages2,zhang2019graphical}, we calculate Recall@50 (R@50), weighted mean AP of relationships ($\text{wmAP}_\text{rel}$), and weighted mean AP of phrases ($\text{wmAP}_\text{phr}$) as evaluation metrics. 
$\text{AP}_\text{rel}$ evaluates the two object bounding boxes, the subject box and the object box, and three labels, the triplets of the subject, the object, and the predicate. 
$\text{AP}_\text{phr}$ evaluates a union bounding box of subject and object and three labels as the same as $\text{AP}_\text{rel}$. 
To reduce the dataset bias in evaluation, we calculate $\text{wmAP}_\text{rel}$ and $\text{wmAP}_\text{phr}$ with weighted average of per-relationship $\text{AP}_\text{phr}$ and $\text{AP}_\text{phr}$, respectively.
The weight of each relationship is calculated by their relative ratios in the validation set. 
The final score $\text{score}_\text{wtd}$ is obtained as $0.2\times \text{R}@50 + 0.4 \times \text{wmAP}_\text{rel} + 0.4 \times \text{wmAP}_\text{phr}$.

\subsection{Implementation details}

As in the previous work~\cite{tang2019learning,li2021bipartite}, we adopt ResNeXt-101-FPN as a backbone network and Faster R-CNN as an object detector. 
The model parameters of the pre-trained object detector are frozen during the training time. 
We use a bi-level sampling~\cite{li2021bipartite} to handle the long-tailed distribution of the datasets. 
The hyperparameters of bi-level sampling are set the same as in \cite{li2021bipartite}.
We set the hyper-parameter $\lambda=0.1$ for the loss function. 
The keeping ratio $\rho$ is set to 70\% for the SGDet setting on both the Visual Genome dataset and the OpenImages v6 dataset in the training.
In the early stages of training, 
 the edge selection model is not reliable, causing instability during training. 
To tackle the issue, we pre-trained the edge selection module for a few thousand iterations  using $\mathcal{L}_{\mathrm{ESM}}$ while freezing all other parameters and then trained the entire \ours except for the object detector.
Complete implementation details are specified in the supplementary material.

\subsection{Comparison with state-of-the-art models}

\begin{table}[!t]
\begin{center}
    \scalebox{0.85}{
    \begin{tabular}{l|P{1,3cm}|P{1.1cm}P{1.1cm}|P{1.3cm}}
    \toprule
    \multirow{2}{*}{Methods} & \multirow{2}{*}{R@50} & \multicolumn{2}{c|}{wmAP} & \multirow{2}{*}{$\text{score}_\text{wtd}$} \\ 
         &  & rel & phr & \\ \midrule
        RelDN~\cite{zhang2019graphical} & 73.1 & 32.2 & 33.4 & 40.9 \\ 
        VCTree~\cite{tang2019learning}&  76.1   & 34.2  & 33.1  &  42.1 \\
        Motifs~\cite{zellers2018neural}& 71.6    & 29.9  & 31.6  &  38.9  \\
        VCTree+TDE~\cite{tang2020unbiased}& 69.3    & 30.7  &  32.8 & 39.3   \\
        GPS-Net~\cite{lin2020gps}& 74.8   & 32.9   & 34.0  & 41.7   \\ 
        GPS-Net$^\dagger$~\cite{lin2020gps}& 74.7 & 32.8 & 33.9 & 41.6 \\
        BGNN$^\dagger$~\cite{li2021bipartite}&  75.0  & 33.5  & 34.2  & 42.1   \\ 
        HL-Net~\cite{lin2022hl} & 76.5 & 35.1 & 34.7 & 43.2 \\
        RU-Net~\cite{lin2022ru} & \bfseries 76.9 & \bfseries 35.4 & 34.9 & \bfseries 43.5 \\
        \ours$^\dagger$ &  75.8 & 34.9 & \bfseries 35.9 & \bfseries 43.5 \\
    \bottomrule
    \end{tabular}
    }
\caption{The scene graph generation performance on OpenImages v6 dataset with graph constraints. $\dagger$ denotes that the bi-level sampling~\cite{li2021bipartite} is applied for the model.
    \label{tb:open_images}}
\end{center}
\end{table}

As shown in Table~\ref{tb:maintable}, on the Visual Genome dataset, \ours outperforms the state-of-the-art models on every setting, PredCls, SGCls and SGDet. 
Especially, \ours 
outperforms the state-of-the-art models by a large margin of 3.9 in mR@100 on the SGDet setting, which is the most realistic and important setting in practice as there is no perfect object detector. 
There are more invalid pairs in the SGDet setting than in other settings since the detected object bounding boxes from the pre-trained object detector includes many background boxes.   
This means that previous work for contextual modeling was most likely distracted by the invalid pairs; thus, \ours shows significant performance improvement on the SGDet setting. 
BGNN~\cite{li2021bipartite} also leverage a scoring function to scale the messages of the invalid edges, 
however, \ours shows better results with our edge selection module to discard invalid object pairs. 
This becomes doable with the quad attention mechanism with edge selection which helps to reduce noise and outliers from invalid pairs more effectively. 
Also, \ours shows the performance improvements by 2.3 and 0.5 on the SGCls and the PredCls settings with mR@100, respectively; 
the more complex and realistic the task, the more noticeable the performance improvement of \ours becomes. 
It shows that \ours, composed of edge selection and quad attention, is appropriate for contextual reasoning to generate scene graphs even in a complex scene.

Also, as shown in Table~\ref{tb:open_images}, \ours achieve competitive results or even outperform compared with the state-of-the-art models on the OpenImages v6 dataset with score$_\mathrm{wtd}$. 
Since there are fewer objects and relationships in the images of the OpenImages v6 dataset than of the Visual Genome, 
the edge selection module seems less effective for the OpenImages v6 dataset. %
As there is a trade-off in recall and mean recall when bi-level sampling is utilized \cite{chen2019knowledge,lu2021context}, 
the result of \ours in Table~\ref{tb:open_images} is a compromised metric for R@50.   %
But still, the R@50 of \ours is still competitive with that from RU-Net~\cite{lin2022ru} and outperforms other recent baselines, and we achieve the best performance in wmAP$_\mathrm{phr}$ by a large margin.   
It shows \ours is effective in improving the scene graph generation performance and also in simple scenes. 

Qualitative visualizations of \ours and more experiments on 
\ours with 1) additional measurement, \eg, recall and non-graph constraint measurement, on Visual Genome, 2) performance with plug-and-play long-tailed recognition techniques and 3) additional qualitative results are given in Supplementary.

\begin{table}[t]

\begin{center}
\scalebox{0.9}{
\begin{tabular}{P{0.8cm}P{0.8cm}P{0.8cm}|P{1.3cm}P{1.3cm}P{1.3cm}}
\toprule
\multicolumn{3}{c}{Variants} & \multicolumn{3}{c}{SGDet} \\
Q & E2E & N2E & \multicolumn{1}{c}{mR@20} & \multicolumn{1}{c}{mR@50} & \multicolumn{1}{c}{mR@100} \\ \midrule
        \multicolumn{3}{c|}{BGNN~\cite{li2021bipartite}}        &     7.49    &      10.31    &  12.46        \\ \midrule
  &  &  & 9.12 & 12.45  & 15.00 \\
 \checkmark &  &  & 9.92  & 13.22 & 15.66 \\
  & \checkmark & \checkmark & 9.84 & 13.04 & 15.60 \\\midrule 
 \checkmark  & \checkmark & \checkmark & 10.57 & 14.12 & 16.47 \\
\bottomrule
\end{tabular}
}
\end{center}
\caption{ The ablation study on model variants on edge selection. We remove the edge selection module for query selection and key-value selection. 
\label{tb:ablation_esm_ox}}

\end{table}

\begin{table}[t]

\begin{center}
\scalebox{0.85}{
\begin{tabular}{P{1.0cm}P{1.0cm}P{1.0cm}|P{1.3cm}P{1.3cm}P{1.3cm}}
\toprule
\multicolumn{3}{c}{Variants} & \multicolumn{3}{c}{SGDet} \\
Q & E2E & N2E & \multicolumn{1}{c}{mR@20} & \multicolumn{1}{c}{mR@50} & \multicolumn{1}{c}{mR@100} \\ \midrule 
 shared & shared & shared & 9.61 & 12.70 & 14.85 \\
 distinct & shared & shared & 9.63 & 12.54 & 14.64 \\\midrule 
 distinct  & distinct & distinct & 10.57 & 14.12 & 16.47 \\
\bottomrule
\end{tabular}
}
\end{center}
\caption{ The ablation study on model variants on edge selection.  We share the edge selection module for query selection and key-value selection. `shared' denotes the edge selection modules share the parameters. 
\label{tb:ablation_esm_shared}}
\end{table}

\subsection{Ablation study}
\subsubsection{Model variants on edge selection}
To verify the effectiveness of the edge selection module, we evaluate the model from which each component of edge selection is removed on the Visual Genome dataset.
As shown in Table~\ref{tb:ablation_esm_ox}, we observe that the quad attention module without the edge selection module shows much lower performance at mR@100 (-8.9\%)  than the full model which has the edge selection module; thus, to select the valid edges is important for the scene graph generation.
On the other hand, the quad attention module without the edge selection module shows 20.4\% higher performance than the BGNN and achieves 15.00 on mR@100.
It shows the effectiveness of the quad attention module itself without the edge selection module. 
We also observe that the query selection is more critical than the key-value selection for the scene graph generation; it shows that selecting what to update is important for the scene graph generation. 

To evaluate the effectiveness of three distinct edge selection modules, we evaluate the models, some of which edge selection modules are shared. 
In Table~\ref{tb:ablation_esm_shared}, $\mathrm{ESM}^{a}$, of which $a\in\{\mathrm{Q}, \mathrm{E2E}, \mathrm{N2E}\}$ are denoted `shared' in the column, share the same parameters. 
We observe that the three fully-distinct edge selection modules boost the scene graph generation performances.
It shows there exist differences between the edges needed to update both features and the edges that need to be updated.

\begin{figure*}[t!] %
\begin{center}
\includegraphics[width=0.99\linewidth]{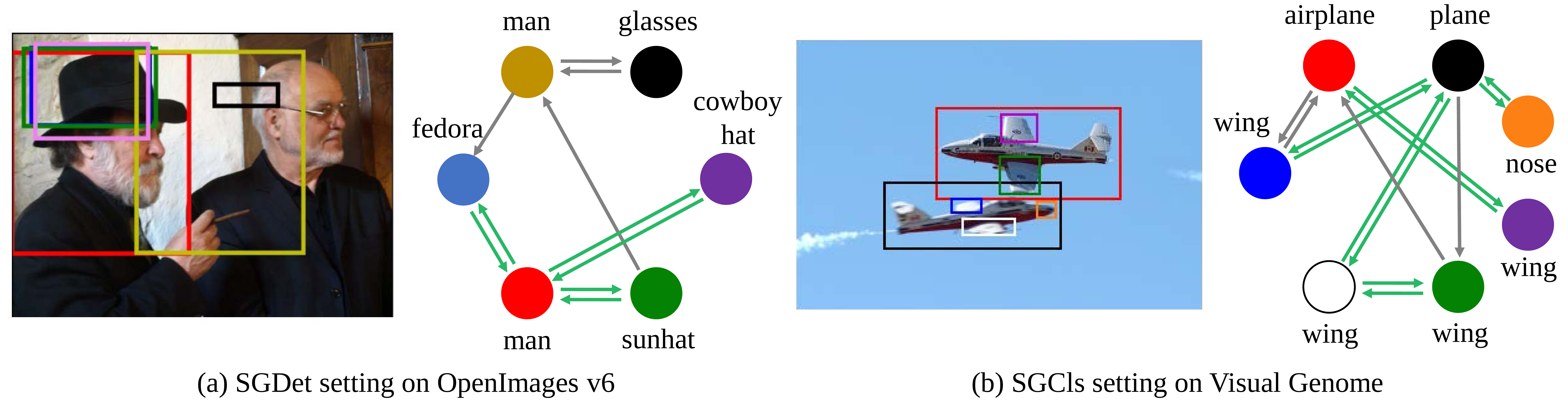}
\end{center}
\caption{Qualitative results for edge selection module $\mathrm{ESM}^\mathrm{Q}$ for query selection. The selected edges after edge selection are drawn in the right graph. The green arrows denote the valid pairs, and the gray arrows denote the invalid pairs. The keeping ratio for the two settings is the same $\rho=35\%.$ All of the valid edges remain, and most of the invalid edges are removed.  }
\label{fig:qual1}
\end{figure*}

\subsubsection{Model variants on quad attention} 

To verify the effectiveness of the quad attention module, we evaluate the model from which each attention is removed on the Visual Genome dataset. 
As shown in Table~\ref{tb:a_model_compo}, the full model with quad attention outperforms the other model variants. 
We also observe that the \ours without updating edges, \ie, edge-to-node and edge-to-edge attentions, performs poorer than the \ours without updating nodes, \ie, node-to-node and node-to-edge attentions. 
It shows that updating edge features with context information is important for context reasoning. 
Especially, \ours without edge-to-edge attention shows worse performance than without edge-to-node attention since edge-to-edge attention, which is neglected in the previous work, can capture high-level information.

\begin{table}[t]
\begin{center}
\scalebox{0.85}{
\begin{tabular}{P{0.65cm}P{0.65cm}P{0.65cm}P{0.65cm}|P{1.3cm}P{1.3cm}P{1.3cm}}
\toprule
\multicolumn{4}{c|}{Method} & \multicolumn{3}{c}{SGDet} \\ 

   N2N & N2E & E2N  &  E2E    &  mR@20   &   mR@50   & mR@100   \\ \midrule
    \checkmark      &     \checkmark   &   &     &   7.02     &     9.74  &     11.57    \\
   \checkmark & \checkmark  &   & \checkmark  & 9.76  & 12.98  & 15.30  \\
    \checkmark &   \checkmark       & \checkmark  &  & 9.70 &     12.27     &     15.03       \\
    &   &  \checkmark & \checkmark     &  9.90 & 13.05  & 15.28  \\
    & \checkmark  & \checkmark  & \checkmark  &  9.77  &  12.93 & 15.42  \\
   \checkmark &   &  \checkmark &  \checkmark &  9.99 & 13.02  & 15.54  \\\midrule
    \checkmark      &   \checkmark & \checkmark &   \checkmark  &   10.57 & 14.12  & 16.47  \\
\bottomrule
\end{tabular}
}
\caption{ The ablation study on model variants on quad attention. N2N, N2E, E2N, E2E denote the node-to-node, node-to-edge, edge-to-node, and edge-to-edge attentions, respectively. 
\label{tb:a_model_compo}}
\end{center}
\end{table}

\subsection{The effect of the edge selection module}
\label{sec:effect_esm}
We applied the edge selection module to the scene graph generation model with message passing methods. 
Since it is not known which pairs of objects have a relationship, the message passing methods use the fully-connected graph in the inference time. 
However, we empirically observe that message passing on fully-connected graph is meaningless or even harmful for the scene graph generation. 
We use three baselines, IMP~\cite{xu2017scene}, BGNN~\cite{li2021bipartite}, and \ours, for ablation study on the Visual Genome dataset.
Table~\ref{tab:exp_message_passing} shows that message passing through fully-connected graph is harmful for the scene graph generation. 
Even though BGNN uses a gating function to scale down the messages from invalid edges, it does not work well.

To further investigate the effect of edge selection, we applied message passing through ground-truth scene graphs to each model. 
In table~\ref{tab:exp_message_passing}, `No' represents that message passing is not used, `Full' and `GT' indicate that message passing is used with the complete graph and the ground-truth scene graph, respectively; `ES' means that the proposed edge selection module is used with message passing. 
As shown in Table~\ref{tab:exp_message_passing}, every model with the message passing through ground truth outperforms state-of-the-art models by a substantial margin, showing that removing the invalid edges is crucial for scene graph generation. 
The edge selection module clearly improves not only the performance of \ours but also that of BGNN, the previous state-of-the-art model. 
It indicates that the edge selection module effectively removes the invalid edges and can be used as a plug-and-play module for message-passing-based scene graph methods. %

\begin{table}[t!]
    \centering\fontsize{9}{11}\selectfont
    \begin{tabular}{c|c|ccc}
    \toprule
         \multirow{2}{*}{model}  &   \multirow{2}{*}{Graph}  & \multicolumn{3}{c}{SGDet} \\
          & &  mR@20 & mR@50 & mR@100 \\ \midrule
         \multirow{2}{*}{IMP~\cite{xu2017scene}} &  No & 4.09 & 5.56 & 6.53 \\ 
                                                 & Full & 2.87 & 4.24 & 5.42 \\\midrule
         \multirow{4}{*}{BGNN~\cite{li2021bipartite}} &  No  & 8.99 & 11.84& 13.56  \\ 
                                                      & Full & 7.49 & 10.31& 12.46  \\
                                                      &  \ds{ES}  & 9.00 & 11.86 & 14.20  \\
                                                      &  GT  & 14.15 & 16.41 & 17.09 \\ \midrule
         \multirow{4}{*}{\ours}                       & No   & 8.68  & 11.52 & 13.99 \\
                                                      & Full & 9.12  & 12.45 & 15.00 \\
                                                      & ES   & 10.57 & 14.12 & 16.47 \\
                                                      & GT   & 17.95 & 19.21 & 19.51 \\
                                                      
                                                    \bottomrule
    \end{tabular}  
    \caption{The ablation study on message passing for the scene graph generation. There are four settings depending on which graphs are used in the message passing: No, Full, ES, and GT. Every result is reproduced with the authors' code.
    \label{tab:exp_message_passing}} 
\end{table}

\subsection{Qualitative results}
Qualitative results for the edge selection module are shown in Fig.~\ref{fig:qual1}. 
As shown in Fig.~\ref{fig:qual1} (a), the object detection module extracts 6 bounding boxes, then the fully-connected graph has 30 edges in total, where only 6 valid edges are in the ground-truth. 
After edge selection with keeping ratio $\rho=35\%$, only 10 edges remain where 6 valid edges all remain. 
It significantly reduces noises from invalid edges. 
The other example in Fig.~\ref{fig:qual1} (b) shows the same tendency. 

\section{Conclusion}

We presented a novel scene graph generation model that predicts a scene graph within an image.  
The method is designed to selectively utilize valid edges using our proposed quad attention module, 
and update the model from the valid edges only. 
The edge selection module effectively %
filters out invalid edges to sparsify a noisy scene graph, and thus it removes uncertainties brought by invalid edges. 
The quad attention module, which is composed of four components --- node-to-node, node-to-edge, edge-to-node, and edge-to-edge attentions --- 
captures the high-level information for accurately predicting relationships among different objects. 
We have shown the effectiveness of the \ours, and each component under various settings was properly validated in the experiments to demonstrate stability.

\paragraph{Acknowledgements.} 
This work was supported by the IITP grants (2021-0-00537: Visual common sense through self-supervised learning for restoration of invisible parts in images (50\%), 2022-0-00959: Few-shot learning of causal inference in vision and language (40\%), and 2019-0-01906: AI graduate school program at POSTECH (10\%)) funded by the Korea government (MSIT).

\makeatletter
\renewcommand \thesection{S}
\renewcommand\thetable{S\@arabic\c@table}
\renewcommand \thefigure{S\@arabic\c@figure}
\renewcommand\thesubsection{\thesection\@arabic\c@subsection}
\setcounter{table}{1}
\setcounter{figure}{1}
\makeatother

\section*{Supplementary Material}

In this supplementary material, we provide additional results and details of our method, Selective Quad Attention Networks (\ours). 

\subsection{Implementation details}
\subsubsection{Code base and GPUs.} 
We implemented \ours using Pytorch~\cite{NEURIPS2019_9015} and some of the official code-base for BGNN~\cite{li2021bipartite}\footnote{https://github.com/SHTUPLUS/PySGG}.
\ours was trained for $\sim$8 hours on 4 RTX 3090 GPUs with batch size 12. 

\subsubsection{Edge selection module.} 
Following \cite{roh2021sparse}, we use simple MLP with 4 linear layers and Layer Normalization~\cite{ba2016layer} with GeLU~\cite{hendrycks2016bridging} activation. 
To capture the global statistics of the edge features $\mathcal{E}=\{f_{ij}\}_{i,j}$, we average half of the output dimensions of the first layer as a global feature $g$: 
\begin{align}
    [h^l_{ij};h^g_{ij}] &= l^1(f_{ij}) \\
    g &= \frac{1}{|\mathcal{E}|} \sum_i \sum_j h^g_{ij},
\end{align}
where $l^1$ is the first layer of the edge selection module and $[\cdot;\cdot]$ is the concatenation operation. The dimensions of the local part $h^l_{ij}$ and the global part $h^g_{ij}$ are the same. 
We concatenate the global feature $g$ with each of the remaining local parts $h^l_{ij}$ and pass into the remaining 3-layer MLP to calculate the relatedness scores $s_{ij}$:
\begin{equation}
    s_{ij}=l^2 ([h^l_{ij};g]),
\end{equation}
where $l^2$ is the remaining 3-layer MLP. 
In order to remove the invalid edges, we choose top-$\rho\%$ highest relatedness score pairs $\mathcal{E}^\rho$ as the valid edges. 

\subsubsection{Training details.} 
To train \ours, we use Stochastic Gradient Descent (SGD) optimizer with a learning rate $10^{-3}$. In the early stages of training, notice that the edge selection model is too naive to select the valid edges to construct feasible scene graphs and therefore causes instability during training. 
To make the training stable, we pre-trained the edge selection module for $2000$ iterations with a learning rate of $10^{-4}$ freezing all other parameters, and then we trained the entire SQUAT without the node detection module.

We use the keeping ratio $\rho=0.7$ and $\rho=0.35$ in training time and inference time, respectively, for all the SGDet settings on the Visual Genome and the Open Images v6 datasets. 
Also, we use the keeping ratio $\rho=0.9$ for the SGCls and the PredCls settings on Visual Genome. 
Since the background proposals do not exist in the SGCls and the PredCls settings, there are fewer invalid edges than in the SGDet setting; thus, we use a smaller keeping ratio. 
We use three quad attention layers for the SGDet setting and two quad attention layers for the SGCls and the PredCls settings.

\begin{table*}[!t]
\begin{center}
\scalebox{0.75} {
\begin{tabular}{l|P{1.7cm}P{1.7cm}P{1.7cm}|P{1.7cm}P{1.7cm}P{1.7cm}|P{1.7cm}P{1.7cm}P{1.7cm}}
\toprule
\multirow{2}{*}{Methods} & \multicolumn{3}{c|}{PredCls} & \multicolumn{3}{c|}{SGCls} & \multicolumn{3}{c}{SGDet} \\ 
     & R@50 / 100 & mR@50/100 & F@50 / 100 & R@50 / 100 & mR@50/100 & F@50 / 100 & R@50 / 100 & mR@50/100 & F@50 / 100\\ \midrule
  IMP+$^\ddagger$~\cite{li2017scene}   &  61.1 / 63.1 & 11.0 / 11.8 & 18.6 / 19.9  &    37.5 / 38.5   &   6.2 / 6.5    &    10.6 / 11.1   &   25.9 / 31.2    &  4.2 / 5.2     &     7.2 / 8.9  \\
  Motifs$^\ddagger$~\cite{zellers2018neural}   &    66.0 / 67.9  &    14.6 / 15.8   &  23.9 / 25.6     &    39.1 / 39.9   &     8.0 / 8.5  &    13.3 / 14.0   &  32.1 / 36.9      &  5.5 / 6.8      &   9.4 / 11.5    \\
  Motifs$^{\ddagger\dagger}$~\cite{zellers2018neural}   &   64.6 / 66.7   &   18.5 / 20.0    &  28.8 / 30.8    &  37.9 / 38.8   &   11.1 / 11.8    &    17.2 / 18.1       &  30.5 / 35.4      &  8.2 / 9.7      &   12.9 / 15.2    \\
  RelDN~\cite{zhang2019graphical}   &    64.8 / 66.7  &    15.8 / 17.2   &  25.4 / 27.3   &  38.1 / 39.3  &  9.3 / 9.6  &   15.0 / 15.4    &    31.4 / 35.9   &     6.0 / 7.3  &    7.2 / 8.9   \\
  VCTree$^{\ddagger}$~\cite{tang2019learning}  &    65.5 / 67.4  &    15.4 / 16.6   &   24.9 / 26.6    &    38.9 / 39.8   &     7.4 / 7.9  &  12.4 / 13.2     &     31.8 / 36.1  &  6.6 / 7.7     &  10.9 / 12.7      \\ 
  MSDN~\cite{li2017scene} &     64.6 / 66.6 &     15.9 / 17.5  &    25.5 / 27.7   & 38.4 / 39.8      &  9.3 / 9.7     & 15.0 / 15.6      &  31.9 / 36.6     &   6.1 / 7.2    &      10.2 / 12.0 \\
  GPS-Net~\cite{lin2020gps} &   65.2 / 67.1   &   15.2 / 16.6    &  24.7 / 26.6     &   39.2 / 37.8    &    8.5 / 9.1   &   14.0 / 14.7    &    31.1 / 35.9   & 6.7 / 8.6      &    11.0 / 13.9   \\
  RU-Net~\cite{lin2022ru} &  \underline{67.7} / \underline{69.6}     &    - / 24.2   &  - / 35.9     &   42.4 / 43.3     &  - / 14.6     &  - / 21.8     &  \underline{32.9} / \underline{37.5}     &  - / 10.8     &  - / 16.8     \\
  HL-Net~\cite{lin2022hl} & 60.7 / 67.0     &  - / 22.8     &   - / 34.0 &   \underline{42.6} / \underline{43.5}     &  - / 13.5   &    - / 20.6   &    \bfseries{33.7} / \bfseries{38.1}    &    - / 9.2   &  - / 14.8      \\
  VCTree-TDE~\cite{tang2020unbiased} &  47.2 / 51.6    &  25.4 / 28.7     &  33.0 / 36.9     &  25.4 / 27.9  &  12.2 / 14.0     &   16.5 / 18.6   &  19.4 / 23.2      &  9.3 / 11.1      &  12.6 / 15.0     \\
  Seq2Seq~\cite{lu2021context}     &   66.4 / 68.5   &  26.1 / 30.5     &  37.5 / 42.2     &    38.3 / 39.0   &     \underline{14.7} / 16.2  &     \underline{21.2} / 22.9 &   30.9 / 34.4    &    9.6 / 12.1   &  14.6 / 17.9     \\
  GPS-Net$^{\ddagger\dagger}$~\cite{lin2020gps}&    64.4 / 66.7  &  19.2 / 21.4     &  29.6 / 32.4     &    37.5 / 38.6   &     11.7 / 12.5   &     17.8 / 18.9  &   27.8 / 32.1     &   7.4 / 9.5     &    11.7 / 14.7   \\
  JMSGG~\cite{xu2021joint}   &  \bfseries 70.8 / \bfseries 71.7    &    24.9 / 28.0   & 36.8 / 40.3      &  \bfseries 43.4 / \bfseries 44.2     &   13.1 / 14.7    &    20.1 / 22.1   &  29.3 / 32.3      &     9.8 / 11.8  &   14.7 / 17.3    \\
  BGNN~\cite{li2021bipartite}$^\dagger$ &  59.2 / 61.3    &  \underline{30.4} / \underline{32.9}      & \bfseries 40.2 / \bfseries 42.8      & 37.4 / 38.5      &  14.3 / \underline{16.5}     &   20.7 / \underline{23.1}    &31.0 / 35.8       & \underline{10.7} / \underline{12.6}      &  \underline{15.9} / \underline{18.7}     \\ \midrule
  \ours{$^\dagger$} (Ours)&   55.7 / 57.9   &  \bfseries 30.9 / \bfseries 33.4    &    \underline{39.7} / \underline{42.4}   &    33.1 / 34.4   & \bfseries 17.5 / \bfseries 18.8      &    \bfseries   22.9 / \bfseries 24.3  &   24.5 / 28.9      &  \bfseries 14.1 / \bfseries 16.5   &  \bfseries  17.9 / \bfseries 21.0   \\ \bottomrule
\end{tabular}
}
\caption{Recall, mean recall and F score of three subtasks on Visual Genome (VG) dataset with graph constraints. $\dagger$ denotes that the bi-level sampling~\cite{li2021bipartite} is applied for the model. $\ddagger$ denotes that the results are reported from the \cite{chen2019knowledge}. Bold numbers indicate the best performances and underlined numbers indicate the second best performances. }
\label{tb:tradeoff}
\end{center}
\end{table*}

\begin{table*}[!t]
\begin{center}
\scalebox{0.80}{
\begin{tabular}{l|P{1.6cm}P{1.6cm}P{1.75cm}|P{1.6cm}P{1.6cm}P{1.75cm}|P{1.6cm}P{1.6cm}P{1.75cm}}
\toprule
\multirow{2}{*}{Methods} & \multicolumn{3}{c|}{PredCls} & \multicolumn{3}{c|}{SGCls} & \multicolumn{3}{c}{SGDet} \\ 

 & ng-mR@20 & ng-mR@50 & ng-mR@100 & ng-mR@20 & ng-mR@50 & ng-mR@100 & ng-mR@20 & ng-mR@50 & ng-mR@100  \\ \midrule
IMP+$^{\ddagger*}$~\cite{li2017scene} &- & 20.3 & 28.9 & - & 12.1 & 16.9 & - & 5.4 & 8.0 \\ 
Frequency$^{\ddagger*}$~\cite{zellers2018neural} & - & 24.8 & 37.3 & - & 13.5 & 19.6 & - & 5.9 & 8.9 \\ 
Motifs$^{\ddagger*}$~\cite{zellers2018neural} & - & 27.5 & 37.9 & - & 15.4 & 20.6 & - & 9.3 & 12.9 \\ 
KERN~\cite{chen2019knowledge} & - & 36.3 & 49.0 & - & 19.8 & 26.2 & - & 11.7 & 16.0 \\ 
GB-NET-$\beta$~\cite{zareian2020bridging} & - & 44.5 & 58.7 & - & 25.6 & 32.1 & - & 11.7 & 16.6 \\ \midrule 
Motifs ~\cite{zellers2018neural} & 19.9&32.8&44.7&11.3&19.0&25.0&7.5&12.5&16.9 \\
VCTree~\cite{tang2019learning}& 21.4& 35.6& 47.8&14.3&23.3&31.4& 7.5&12.5&16.7\\
VCTree-TDE~\cite{tang2020unbiased}& 20.9 & 32.4 & 41.5& 12.4& 19.1 & 25.5 & 7.8&11.5& 15.2 \\
GPS-Net$^{\dagger*}$~\cite{lin2020gps} & 29.4& 45.4&57.1&8.3  &15.9&23.1 &7.9 &12.1&16.7\\
\ours$^{\dagger}$ & \bfseries 31.8 & \bfseries 46.0 &\bfseries 57.8 & \bfseries{18.7}   & \bfseries{27.1}  & \bfseries{32.6}  &   \bfseries{12.1}   &     \bfseries 17.9   &   \bfseries  22.5  \\
\bottomrule
\end{tabular}
}
\end{center}
\caption{The scene graph generation performance of three subtasks on the Visual Genome (VG) dataset without graph constraints. $\dagger$ denotes that the bi-level sampling~\cite{li2021bipartite} is applied for the model. $*$ denotes that the model is reproduced with the authors' code. $\ddagger$ denotes that the results are reported from the \cite{chen2019knowledge}. Models in the first group use pre-trained Faster R-CNN with VGG16 backbone. Bold numbers indicate the best performances.}
\label{tb:ngmR}
\end{table*}

\subsection{Additional evaluations on Visual Genome}

\subsubsection{Trade-off between recall and mean recall} 
Since the Visual Genome dataset\footnote{The most frequent entity class is 35 times larger than the least frequent one and the most frequent predicate class is 8,000 times larger than the least frequent one.} 
has extremely long-tailed distribution, there is the trade-off between recall and mean recall~\cite{tang2020unbiased,lu2021context}.
To evaluate various trade-offs of the scene graph generation methods, Zhang \etal~\cite{zhang2022fine} propose the F@$K$ measure, the harmonic mean of recall and mean-recall, recently. 
Table~\ref{tb:tradeoff} shows the R@50/100, mR@50/100, and F@50/100 on the Visual Genome dataset.
\ours outperforms all of the state-of-the-art methods at F@50/100 measurements. 
It shows that although the recall of \ours degrades, the trade-off between the recall and the mean recall is the best in the state-of-the-art methods.

\begin{table}[t]
\scalebox{0.75}{
\begin{tabular}{c|ccc|ccc}
\toprule
    model & head & body & tail & mR@100 & R@100 & F@100 \\ \midrule
    VCTree-TDE~\cite{tang2020unbiased} &  24.5 & 13.9 & 0.1  & 11.1 & 23.2 & 15.0 \\
    GPSNet$^{\dagger}$~\cite{lin2020gps} & 30.4 & 8.5 & 3.8 & 9.5 & 32.1 & 14.7 \\
    BGNN$^{\dagger}$~\cite{li2021bipartite} & \bfseries 34.0 & 12.9 & 6.0 & 12.6 & \bfseries 35.8 & 18.6 \\ \midrule
    \ours$^{\dagger}$ (Ours) & 29.5 & \bfseries 16.4 & \bfseries 12.4 & \bfseries 16.5 & 28.9 &  \bfseries 21.0 \\ \bottomrule
\end{tabular}
}
\caption{mR@100 on the SGDet setting for head, body, and tail classes. $\dagger$ denotes that the bi-level sampling is applied on the model to achieve these results. Bold numbers indicate the best performances.}
\label{tb:group}
\end{table}

\subsubsection{Mean recall with no-graph constraints}
Following \cite{newell2017pixels,zellers2018neural}, we also evaluate \ours without the graph constraint, \ie, each edge can have multiple relationships. 
For each edge, while mR$@K$ evaluates only one predicate with the highest score, ng-mR$@K$ evaluates all 50 predicates. 
As shown in Table.~\ref{tb:ngmR}, on the Visual Genome dataset, \ours outperforms the state-of-the-art models. 
Especially, \ours outperforms the state-of-the-art models by a large margin of ng-mR$@K$ on the SGDet settings as it does in the evaluation of mR$@K$.

\subsubsection{Recall for head, body, and tail classes}
Following~\cite{li2021bipartite}, we split the relationship classes into three sets according to the number of relationship instances: head (more than 10k), body (0.5k$\sim$10k), and tail (less than 0.5k) classes. Table~\ref{tb:group} shows the mR@100 for each group. 
\ours outperforms the state-of-the-art methods for body and tail classes by a large margin. 
Especially for the tail classes, \ours achieves twice mR@100 as that of BGNN. 
It shows that the scene graphs from \ours have more meaningful predicates, \ie, tail classes such as `walking in', instead of general predicates, \ie, head classes such as `on'.

\subsubsection{Recall on simple, moderate, and complex scenes}
As shown in Tables~\ref{tb:maintable} and~\ref{tb:open_images} in the main paper, the SQUAT shows exceptionally high performance on the most complicated task, \ie, SGDet, and the most complex dataset, \ie, Visual Genome. 
Furthermore, to analyze the performance on the complexity of the scene, we divide the image sets in the Visual Genome into three disjoint sets according to the number of objects in the scene: simple ($\leq9$), moderate ($10\sim16$), and complex ($\geq 17$). 
As shown in Table~\ref{tab:reb_complex}, the SQUAT shows a higher performance gain on the more complex images; the SQUAT is more effective for realistic and complex scenes. 

\begin{table}[t]
    \centering\fontsize{9}{10}\selectfont
    \begin{tabular}{c|ccc|c}
    \toprule
         model  & simple & moderate & complex & mR@100  \\ \midrule
         BGNN [18]   &  15.52   &  12.71   &  9.87   &  12.46     \\
         SQUAT       &  19.54   &  16.80   &  13.28   &  16.47     \\ \midrule 
         Gain (\%)   &  25.90   &  32.18    &  34.55    &  32.18      \\ \bottomrule
    \end{tabular}
    \caption{mR@100 on the simple, moderate, and complex sets. 
    \label{tab:reb_complex}}
\end{table}

\subsection{\ours with off-the-shelf method}
To reduce the biases of the scene graph generation datasets, many off-the-shelf methods~\cite{suhail2021energy,yan2020pcpl,Yao_2021_ICCV,Desai_2021_ICCV,Guo_2021_ICCV,Knyazev_2021_ICCV,chiou2021recovering,knyazev2020graph,wang2020tackling,zhang2022fine,deng2022hierarchical} are proposed. 
For a fair comparison, we do not compare the off-the-shelf methods with \ours in the main paper. 
We applied Internal and External Data Transfer (IETrans) and reweighting (Rwt)~\cite{zhang2022fine}, which are the state-of-the-art off-the-shelf learning methods for scene graph generation, to the \ours. 
For efficiency, we only report a model with the best performance for each off-the-shelf method. 
As shown in Table~\ref{tab:supp_bplsa}, without careful hyper-parameter search, SQUAT+IETrans+Rwt model outperforms  VCTree+IETrans+Rwt model and outperforms other off-the-shelf methods with Motifs~\cite{zellers2018neural}, Transformer~\cite{tang2020unbiased}, and VCTree~\cite{tang2019learning}. 
It shows that 
other off-the-shelf learning methods can be adopted for \ours to improve its performance. 

\begin{table}[t!]
    \centering\fontsize{9}{11}\selectfont
    \begin{tabular}{l|ccc}
    \toprule
         \multirow{2}{*}{model}  & \multicolumn{3}{c}{SGDet} \\
           &  mR@20 & mR@50 & mR@100 \\ \midrule
         VCTree~\cite{tang2019learning} &   4.9 &  6.6  &  7.7  \\
         VCTree+TDE~\cite{tang2020unbiased} & 6.3 & 8.6 & 10.3 \\
         VCTree+PCPL $^\ddagger$~\cite{yan2020pcpl} & 8.1 & 10.8 & 12.6 \\
         VCTree+DLFE~\cite{chiou2021recovering} & 8.6 & 11.8 & 13.8 \\
         VCTree+TDE+EBM~\cite{suhail2021energy} & 7.1 & 9.69 & 11.6 \\ 
         Transformer+BPL+SA~\cite{Guo_2021_ICCV} & 10.7 & 13.5 & 15.6 \\
         Transformer+HML~\cite{deng2022hierarchical} &  11.4 & 15.0 &  17.7 \\
         GPSNet+IETrans+Rwt~\cite{zhang2022fine} & - & 16.2 & 18.8 \\ 
         \midrule
         \ours+IETrans+Rwt~\cite{zhang2022fine} & \bfseries 12.0 & \bfseries 16.3 & \bfseries 19.1 \\
    \bottomrule
    \end{tabular}  
    \caption{The ablation study with the off-the-shelf learning methods on Visual Genome (VG) dataset with graph constraint. $\ddagger$ denotes that the results are reported from the~\cite{chiou2021recovering}. The other results are from each of the original papers.} 
    \label{tab:supp_bplsa}
\end{table}

\subsection{Additional qualitative results}

\begin{figure*}[t!] %
\begin{center}
\includegraphics[width=0.99\linewidth]{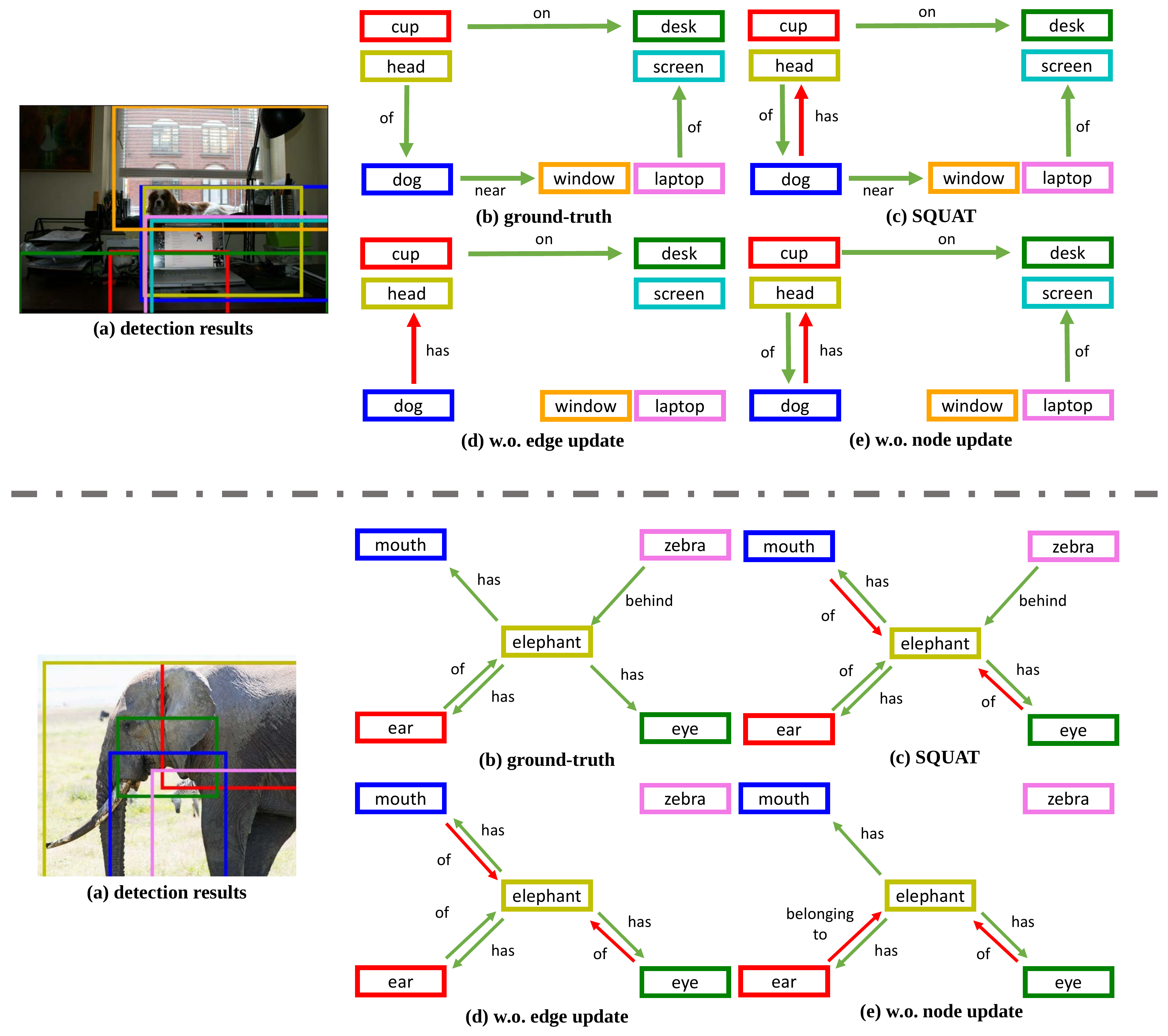}
\end{center}
\caption{The qualitative results for \ours. (a) The detection results from pre-trained Faster R-CNN~\cite{ren2015faster}. (b) The ground-truth scene graph. (c) The results from full \ours. (d) The results from \ours without edge update, \ie, the edge-to-edge and the edge-to-node attentions. (e) The results from \ours without node update, \ie, the node-to-edge and the node-to-node attentions. 
Full \ours shows more informative scene graphs than the other ablated models. The green arrows denote the true positives and the red arrows denote the false positives.}
\label{fig:supp_qual1}
\end{figure*}

\begin{figure*}[t!] %
\begin{center}
\includegraphics[width=0.99\linewidth]{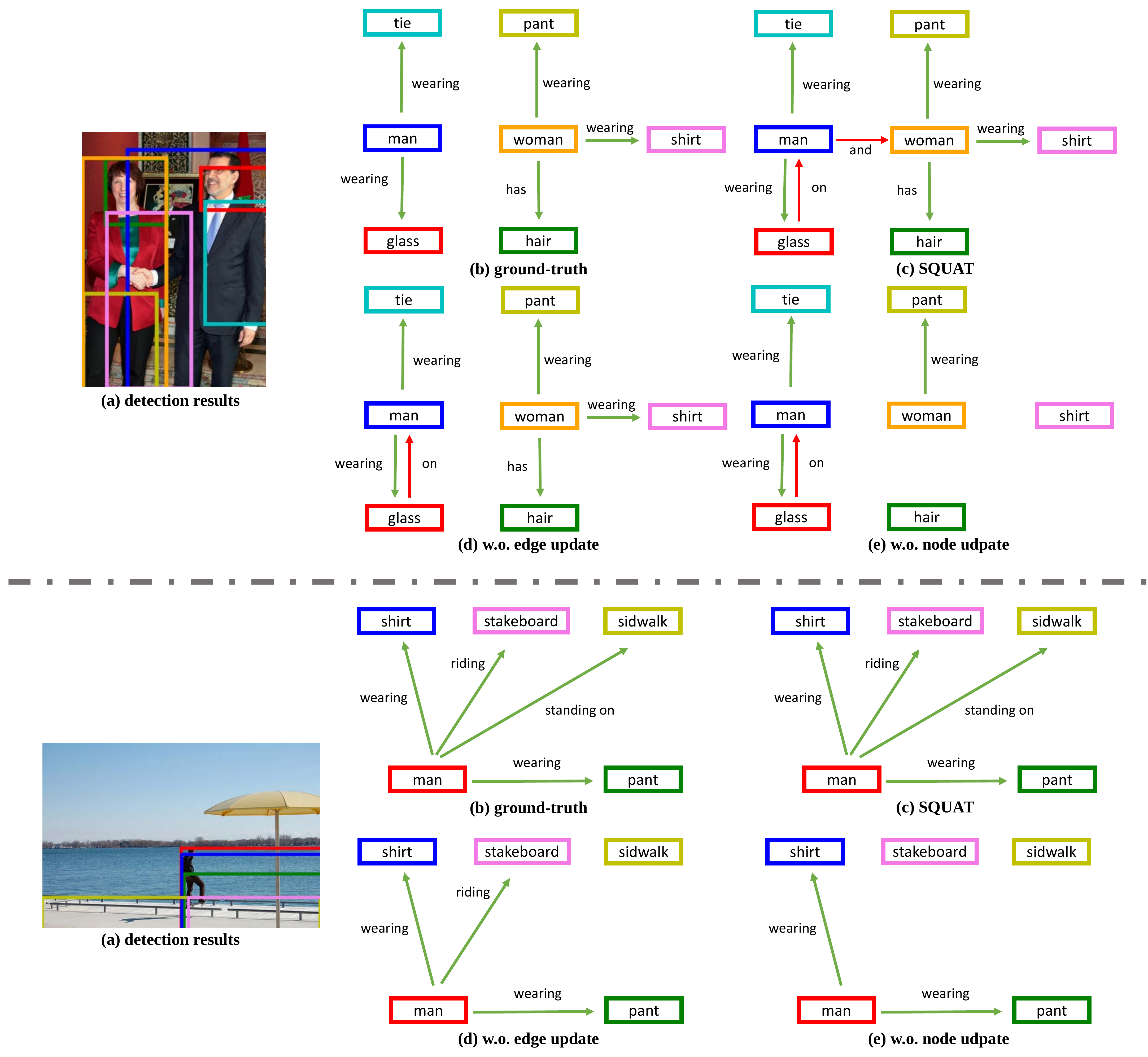}
\end{center}
\caption{The qualitative results for \ours. (a) The detection results from pre-trained Faster R-CNN~\cite{ren2015faster}. (b) The ground-truth scene graph. (c) The results from full \ours. (d) The results from \ours without edge update, \ie, the edge-to-edge and the edge-to-node attentions. (e) The results from \ours without node update, \ie, the node-to-edge and the node-to-node attentions. 
Full \ours shows more informative scene graphs than the other ablated models. The green arrows denote the true positives and the red arrows denote the false positives.}
\label{fig:supp_qual2}
\end{figure*}

In Fig.~\ref{fig:supp_qual1} and \ref{fig:supp_qual2}, we show the qualitative results for \ours model. We also compare the results of \ours with the results from ablated models: model without node updates and model without edge updates. 
The full \ours model shows the most informative scene graph compared to the other ablated models. 
There are some false positives, such as (`mouth of elephant', `eye of elephant') in Fig.~\ref{fig:supp_qual1} bottom and (`glasses on man', `man and woman') in Fig.~\ref{fig:supp_qual2} top, however, such errors are often caused by 
the incompleteness of the dataset, and hence it can be seen as a true positive.  
`dog near window' in Fig.~\ref{fig:supp_qual1} top, `zebra behind elephant' in Fig.~\ref{fig:supp_qual1} bottom, and `man standing on sidewalk' in Fig.~\ref{fig:supp_qual2} bottom are predicted by only the full \ours model. 
It shows that quad attention modules can capture more informative contextual information. 

\begin{figure*}[t!] %
\begin{center}
\includegraphics[width=0.99\linewidth]{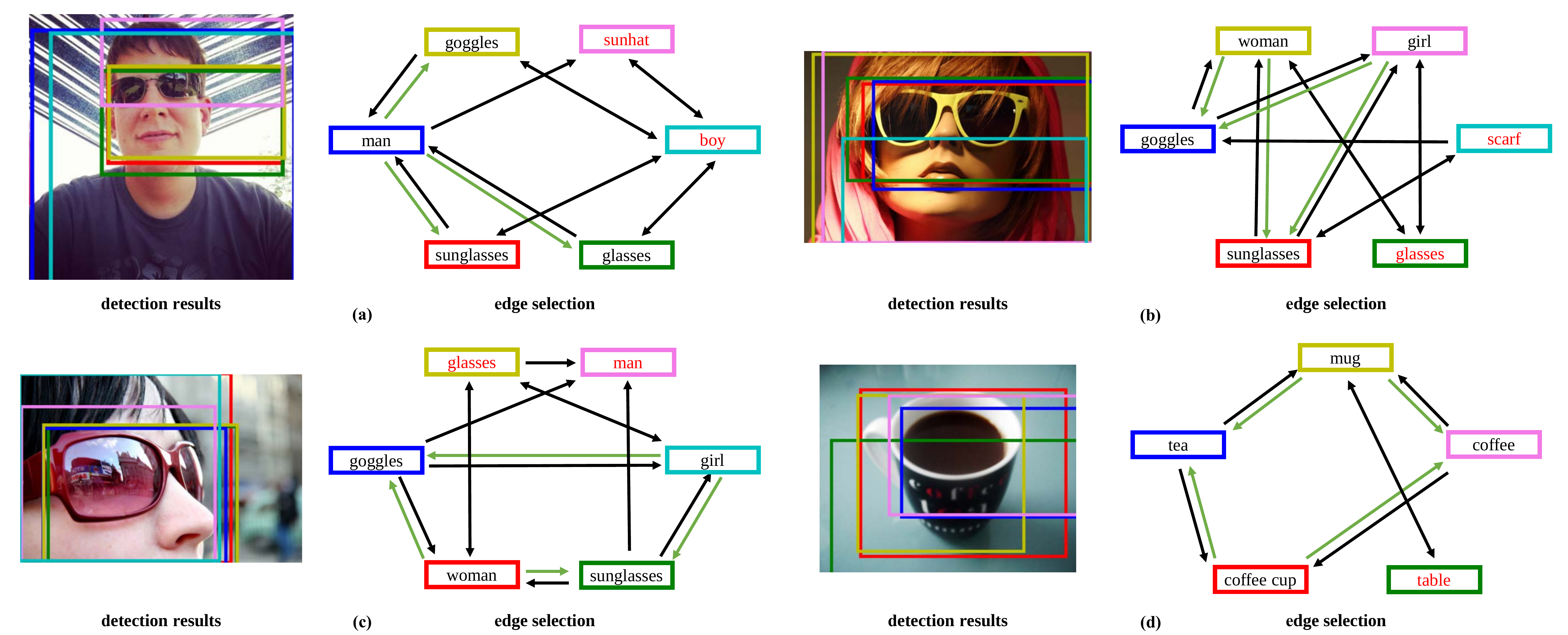}
\end{center}
\caption{The qualitative results for the edge selection module on the Open Images v6 dataset. The graph denotes the results of the $\mathrm{ESM}^\mathrm{Q}$ and the green arrows denote the valid edges. The boxes with the red class denote the incorrect prediction or the background.}
\label{fig:supp_qual_edge}
\end{figure*}

In Fig.~\ref{fig:supp_qual_edge}, we show the qualitative results for the edge selection module in \ours. 
The edge selection module successfully selects the valid edges.
In particular, the edge selection module removes the edges between the background and the foreground, \textit{e.g.}, most of the edges of `sunhat' and `scarf' are removed in Fig.~\ref{fig:supp_qual_edge} (a) and (b), respectively. 
Also, the edges between the boxes which denote the same objects are removed.
For example, the edges of (`tea', `coffee') and (`mug', `coffee cup') are removed in Fig.~\ref{fig:supp_qual_edge} (d). 
It shows that the edge selection module successfully removed invalid edges and helps the informative message passing in the quad attention module. 

\clearpage
\clearpage

{\small
\bibliographystyle{ieee_fullname}
\bibliography{egbib}
}

\end{document}